\documentclass[fleqn,10pt]{JLA_article}

\usepackage[english]{babel}
\usepackage[table]{xcolor}
\usepackage{url}
\usepackage{hyperref}
\hypersetup{
  hidelinks,
  colorlinks,
  breaklinks    = true,
  urlcolor      = color2,
  citecolor     = color1,
  linkcolor     = color1,
  bookmarksopen = false,
  pdftitle      = {An Auditable Policy-Simulation Framework for Student Dropout in Intervention-Free Data},
  pdfauthor     = {Rafael da Silva, Jeff Eicher, Gregory Longo}
}
\usepackage{graphicx}
\usepackage{subcaption}
\usepackage{amsmath}
\usepackage{amssymb}
\usepackage{array}
\usepackage{booktabs}
\usepackage{amsthm}
\usepackage{algorithm}
\usepackage{algpseudocode}
\usepackage{xurl}
\usepackage{enumitem}

\graphicspath{{../paper4/images/}}

\newtheorem{proposition}{Proposition}
\theoremstyle{definition}

\theoremstyle{plain}

\newcommand{\artifact}[1]{\nolinkurl{#1}}
\PaperTitle{An Auditable Policy-Simulation Framework for Student Dropout in Intervention-Free Data}

\Authors{Rafael da Silva\textsuperscript{1}*, Jeff Eicher\textsuperscript{1}, Gregory Longo\textsuperscript{1}}
\affiliation{\textsuperscript{1}\textit{Applied Data Science Program, Eastern University, St.\ Davids, PA, USA.
  \{rafael.dasilva, jeff.eicher, gregory.longo\}@eastern.edu}}


\Submitted{}
\Accepted{}
\Published{}
\Volume{}
\Number{}
\Pages{}
\Doi{}

\Notesname{Notes for Practice}

\note{Early-warning systems in higher education typically identify \emph{who} is at risk without revealing \emph{when} risk peaks during the course. The discrete-time hazard framework presented here translates weekly LMS engagement into time-indexed risk trajectories that advisors can inspect and act on weekly---without model retraining---directly addressing the timing gap between risk detection and support delivery.}

\note{A scenario-based policy simulation layer enables institutions to contrast projected student survival under alternative hypothetical intervention rules (different recency thresholds, active windows, and intensity assumptions), yielding a fully parameterised, auditable operator specification that can be re-run and inspected without access to the training data or code.}

\note{A subgroup equity monitor quantifies whether a given trigger rule differentially changes survival gaps across demographic groups (gender, disability status), with bootstrap-estimated uncertainty. Practitioners can embed this lightweight diagnostic in routine learning analytics dashboards to track whether an intervention policy amplifies or reduces existing outcome disparities.}

\Abstract{%
Institutions hold rich observational learning data but no record of interventions, leaving them unable to tell from that data alone \emph{when}---or \emph{whether}---to act on dropout risk. We address this gap with a temporal modeling framework whose contribution is an \emph{auditable} policy-simulation operator. A discrete-time penalized-logistic hazard backbone estimates weekly withdrawal risk over person-period records; an exportable, re-runnable operator then links those hazards to explicit intervention-timing rules and reports how alternative support scenarios change projected survival and subgroup gaps---without model retraining. The framework is evaluated on the Open University Learning Analytics Dataset (OULAD; $N{=}32{,}593$ enrollments). Weekly hazard discrimination is $\text{AUC}_{row}{=}0.84$ with tight aggregate calibration ($\text{ECE}_{15}{=}0.0012$); the figure that governs enrollment-level decisions is the horizon discrimination $\text{AUC}_{IPCW}{=}0.77$, recovered by a mean-hazard recalibration that resolves a product-form rank inversion we characterize formally. Discrimination transfers to five held-out course runs ($\text{AUC}_{row}{>}0.76$). Under an explicit trigger/schedule contract the operator yields a monotone, inspectable dose--response of survival contrasts $\Delta S$ with bootstrap uncertainty; because product-form survival is monotone in each hazard, a positive contrast under a hazard-reducing scenario is definitional, so the reportable content is the operator's behavior, not an effect estimate. A subgroup monitor returns directionally stable but negligible gap changes ($|\Delta\text{Gap}|\approx 5\times 10^{-4}$), evidencing that the tooling is operational rather than a substantive equity effect. All policy and subgroup quantities are model-implied structural contrasts, not causal effects. The deliverable is a reproducible, retraining-free tool that advisors can run weekly.
}

\begin{document}

\flushbottom
\maketitle
\thispagestyle{fancy}


\section{Introduction}
\addcontentsline{toc}{section}{Introduction}

In the United States alone, roughly 30\% of undergraduates withdraw after their first year---costing over \$9~billion in unrealized public investment and foreclosing social mobility pathways~\cite{Schneider2010FirstYearAttrition}---yet most early-warning systems tell counselors \emph{who} is at risk without revealing \emph{when} risk peaks, leaving institutions unable to time a support nudge to the onset of disengagement. Rather than conceptualizing dropout as a static outcome, we model withdrawal as a dynamic process unfolding over academic time.

A large body of work has studied dropout prediction using machine-learning models built from demographic, academic, and engagement signals. However, many approaches still emphasize static, single-time risk estimates~\cite{Prenkaj2021SurveyMLDropout}. In practice, moving from prediction to action requires (i)~expressing risk in temporal terms, (ii)~translating temporal risk into auditable decision rules and timing~\cite{MartinezCarrascal2023SurvivalOULAD}, and (iii)~comparing alternative support scenarios, as informed by the OnTask approach~\cite{Pardo2018OnTask}. These three requirements correspond directly to RQ1 (temporal risk modeling), RQ2 (policy simulation and trigger-schedule design), and RQ3 (subgroup-sensitive scenario analysis).

\noindent\textit{From prediction to policy.}
Static risk scores primarily indicate \emph{who} is at risk, but operational support also depends on \emph{when} risk increases and which decision rule triggers action. Once risk is expressed as a time-indexed hazard trajectory, the next step is to contrast model-implied survival trajectories across alternative hypothetical scenarios, enabling structural evaluation of competing support policies~\cite{Keil2014ParametricGFormula,Dickerman2022CounterfactualRisks}.

\noindent\textit{Conceptual framework.}
Our framing places this work deliberately on the model-implied side of a recognized methodological boundary: \emph{counterfactual prediction without causal identification}~\cite{DickermanHernan2020CounterfactualPrediction}. \citeA{DickermanHernan2020CounterfactualPrediction} distinguish factual prediction from counterfactual prediction and treat ``what-if'' quantities under specified policies as a legitimate prediction task, while noting that their validity rests on the same unverifiable assumptions as causal inference. \citeA{Prosperi2020CausalCounterfactual} caution, in turn, that data-driven predictive models do not on their own carry a causal interpretation. We do not cross that frontier: $\Delta S(t)$ and $\Delta Gap(T)$ are structural, model-implied contrasts, not causally identified effect estimates. The contribution is the auditable simulation operator and the gap diagnostic, not the identification of intervention effects.

We operationalize dropout as administrative withdrawal from a course run with an observed unregistration time, enabling a time-to-event representation in discrete academic weeks. The predictive backbone is a discrete-time hazard framework fit on weekly person--period observations, with calibration to yield interpretable hazard probabilities over time.

Beyond prediction, the framework incorporates a simulation layer for counterfactual policy analysis, broadly consistent with structural comparison frameworks in causal inference~\cite{Keil2014ParametricGFormula,Wen2020ParametricGFormula}. Under explicit user-defined parameters, model-implied hazards are scaled to represent a hypothetical support scenario, enabling comparison of survival trajectories across structural scenarios. Predicted hazards are interpreted as candidate warning windows for structural sensitivity analysis of intervention timing and magnitude along each enrollment trajectory~\cite{Dickerman2022CounterfactualRisks,Wen2020ParametricGFormula}.

The framework is applied in a subgroup-sensitive way for groups defined by observable characteristics. In the reported run, the subgroup signal is small in absolute magnitude ($|\Delta\text{Gap}| \approx 5 \times 10^{-4}$) but directionally stable across bootstrap resamples; the framework is intended as a diagnostic equity-monitoring tool, not as evidence of large intervention effects.

This study uses the Open University Learning Analytics Dataset (OULAD), which combines time-stamped VLE interaction logs, assessment information, and administrative records~\cite{Kuzilek2017OULAD}. The paper evaluates the framework through three research questions (RQ1--RQ3) corresponding to: (i)~temporal risk modeling and calibration; (ii)~scenario-based policy simulation; and (iii)~subgroup validation via gap analysis.

\section{Background}

Student dropout is consequential for both learners and institutions, and its dynamics often become visible through changes in engagement across an academic term.

\subsection{Why Dropout Matters: Impacts and Scale}

Student dropout has far-reaching consequences at the individual, family, institutional, and societal levels~\cite{Kamissa2020DroppingOut,Schneider2010FirstYearAttrition,Nadeem2021DropoutUniversityRevenue}. In the United States, approximately 30\% of undergraduates drop out after their first year, resulting in over nine billion dollars in public expenditures over five years~\cite{Schneider2010FirstYearAttrition}. Globally, substantial dropout persists across systems, as documented in socio-economic reviews~\cite{Aina2022DeterminantsDropout}, and course-level dropout may further increase the risk of program withdrawal, particularly among marginalized students~\cite{McKinney2019CourseDropping,Gicheva2025PersistenceIntroCourse}.

\subsection{Structural Shifts in Delivery: From Distance to Online}

Although online learning appears recent, distance education has historical roots in correspondence-based instruction~\cite{Kentnor2015DistanceEducation}. The COVID-19 pandemic accelerated the shift to online instruction as a core mode~\cite{Bartolic2022COVIDTeaching,Bryson2020RapidAdoption,GarciaMorales2021COVIDTransformation,Bond2021EmergencyRemoteTeaching}, and Learning Management Systems consequently became important infrastructures for teaching and engagement~\cite{Turnbull2021TransitioningELearning}. Dropout in online higher education reflects intersecting academic, institutional, and personal barriers~\cite{Rahmani2024OnlineDropoutReview,Rovai2003DistanceEducationPersistence,Lee2013OnlineDropoutSelfRegulation,BarraganMoreno2024ComplexitiesDropout}. Although machine-learning models can achieve high predictive accuracy, challenges persist in data heterogeneity, reproducibility, and temporal coverage~\cite{Albreiki2021MLPerformanceReview}.

\paragraph*{Survival Analysis in Educational Prediction.}
Models such as DeepHit~\cite{LeeDeepHit2018} and Deep Survival Machines~\cite{NagpalDSM2021} offer different architectural trade-offs in temporal dependency modelling. The present study uses a discrete-time logistic hazard backbone, the standard approach when event times are recorded in coarse units such as weeks~\cite{Ellison2022DiscreteTimeEHA,SchmidBerger2020CompetingRisks,SingerWillett1993DiscreteTimeSurvival,Allison1982DiscreteTimeMethods}. Within OULAD specifically, prior work has applied temporal methods to dropout risk~\cite{Hlosta2017OurobosOULAD,MartinezCarrascal2023SurvivalOULAD,Rizvi2019DemographicsOULAD}, but none combines policy simulation, bootstrap uncertainty quantification, and a temporal hazard framework with an intervention-timing contract.

\subsection{Intervention Practice Gaps}

In many higher-education settings, intervention practices are deployed in an ad hoc manner and insufficiently documented to support systematic evaluation~\cite{Delnoij2020NonCompletionReview,Kizilcec2020ScalingInterventions,Sonderlund2018LAInterventions,WongLi2019LAInterventionReview}. Routine observational data typically lack the granularity required to support credible causal evaluation of intervention effects~\cite{Weidlich2022CausalBiasLA}. Although randomized experiments can in principle address these identification gaps, their implementation in live academic environments is frequently constrained by operational considerations~\cite{Harackiewicz2018TargetedIntervention,StylesTorgerson2018RCTsEducation}. In this setting, scenario-based structural comparison provides a disciplined way to study how a fitted temporal model behaves under explicit intervention rules~\cite{Keogh2023CounterfactualPrediction,Boyer2023CounterfactualPredictionModels}.

\subsection{Fairness, Bias, and Group Performance Gaps}

Machine-learning models for dropout prediction carry potential for disparate impact across demographic subgroups~\cite{PessachShmueli2022FairnessReview,Opoku2025AccuracyFairness}. Fairness-aware analysis has moved toward explicit gap-quantification approaches~\cite{PessachShmueli2022FairnessReview,Kizilcec2020ScalingInterventions}, though fairness criteria can conflict: equalizing predictive accuracy and calibration parity are not always simultaneously attainable~\cite{PessachShmueli2022FairnessReview}. The present study contributes by quantifying, for each simulated policy scenario, how the modeled survival gap between gender and disability subgroups changes under the same intervention rule, with bootstrap-estimated uncertainty.

Collectively, the four themes above reveal interconnected gaps addressed by the present framework: (i)~most dropout models answer \textit{who} is at risk but not \textit{when} to act (RQ1); (ii)~prior OULAD-based temporal analyses lack policy simulation and uncertainty quantification (RQ2); (iii)~intervention evaluation is hampered by undocumented, ad hoc support practices; and (iv)~fairness analyses of prediction-derived policies rarely include bootstrap-estimated uncertainty of gap changes (RQ3).

\section{Methodology}

The framework maps temporally indexed enrollment data to weekly hazard estimates, structural contrasts, and subgroup diagnostics. It combines person-period construction, leakage-aware hazard modeling, policy simulation, and subgroup analysis under explicit inferential limits.

\begin{figure*}[ht]
  \centering
  \includegraphics[width=\textwidth]{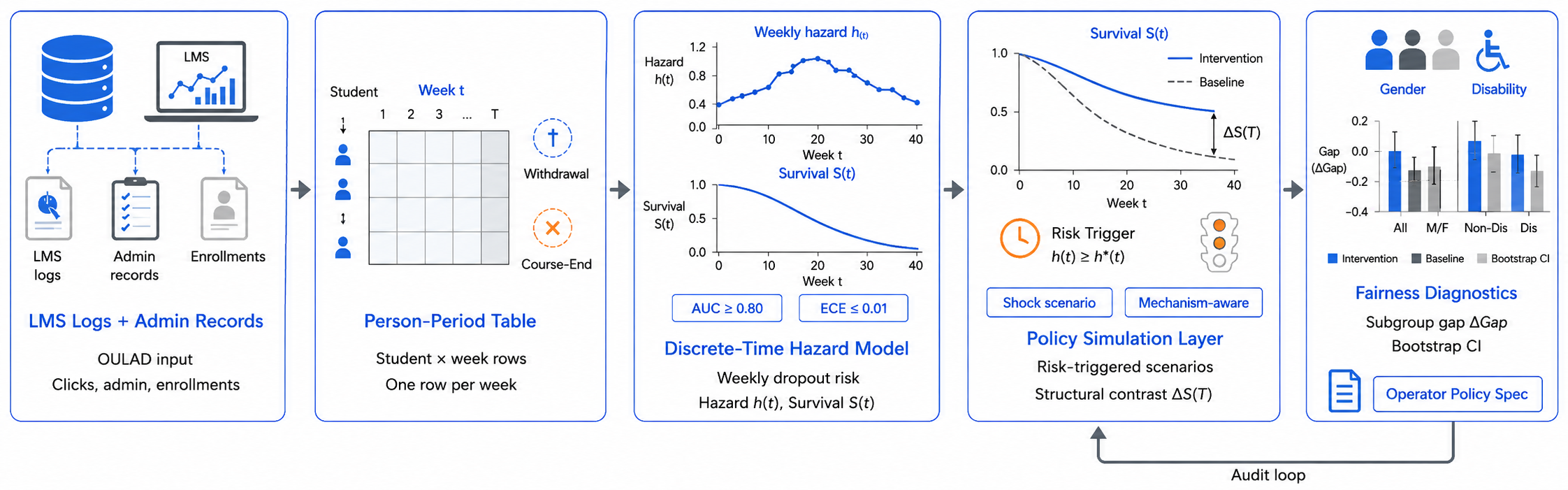}
  \caption{Framework overview: from raw LMS interaction logs and administrative records (\textbf{Input}) through person-period construction, discrete-time hazard modeling (RQ1), scenario-based policy simulation (RQ2), to subgroup fairness diagnostics and exportable policy specification (RQ3).}
  \label{fig:framework_overview}
\end{figure*}

\subsection{Motivation, Contributions, and Research Questions}

\paragraph*{Contributions.}
This study makes four principal contributions:
\begin{enumerate}[noitemsep]
  \item \textbf{Temporal hazard framework}: a discrete-time hazard pipeline for estimating weekly survival trajectories from person-period LMS records, with calibration and leakage controls.
  \item \textbf{Policy simulation layer}: a scenario-based simulation operator that compares \textit{shock} (model-level perturbation) and \textit{covariate-shift} (covariate update) scenarios under a common trigger-and-schedule contract. The contribution is the \emph{auditability} of this operator---its trigger rule, schedule, and intensity are exposed as an inspectable, re-runnable specification (Algorithm~\ref{alg:pipeline})---rather than the sign of the resulting contrast: because product-form survival is monotone in each weekly hazard, a hazard-reducing scenario raises $\Delta S$ by construction, so a positive contrast is not in itself an empirical finding. Unlike prior scenario-based survival work (e.g.,~\cite{Keogh2023CounterfactualPrediction,Boyer2023CounterfactualPredictionModels}), this layer is fully parameterized and exports an executable operator specification rather than a fixed result.
  \item \textbf{Subgroup gap analysis}: a fairness diagnostic layer that quantifies how the same policy changes survival gaps across observable groups (gender, disability), with bootstrap uncertainty.
  \item \textbf{Reproducible artifact pipeline}: an end-to-end implementation that exports all operator contracts, horizon tables, bootstrap draws, and comparison artifacts, enabling external audit and re-use~\cite{RafaRepoTCMStudentDropout}.
\end{enumerate}

\paragraph*{Research Questions.}
\begin{description}
  \item[\textbf{RQ1}] Can a discrete-time hazard model calibrated on person-period LMS data produce weekly withdrawal-risk estimates that are sufficiently discriminative ($AUC_{row}\ge 0.80$) and calibrated ($ECE\le 0.01$) to support time-indexed decision-making?
  \item[\textbf{RQ2}] What structural survival contrast $\Delta S(T)$ emerges between a baseline and a simulated support scenario under an explicit trigger-and-schedule policy contract, and how does it vary with scenario intensity?
  \item[\textbf{RQ3}] Does the same policy produce a differential change in survival gaps across gender and disability subgroups, and how large and directionally stable is that change under bootstrap uncertainty?
\end{description}

\paragraph*{Epistemic status.}\label{para:interp_contract}
The framework combines three kinds of object, deliberately kept distinct: (i)~the \emph{predictive backbone} (discrete-time logistic hazard, Eq.~\ref{eq:hazard_weekly}) is a standard, literature-validated estimator~\cite{SingerWillett1993DiscreteTimeSurvival,Allison1982DiscreteTimeMethods}; (ii)~the \emph{simulation operator} (Eq.~\ref{eq:delta_s}; Algorithm~\ref{alg:pipeline}) is a deterministic, reproducible transformation of fitted hazards under an explicit policy contract---which makes \emph{no} causal-identification claim; (iii)~Proposition~\ref{prop:rank_inversion} is the single derived result.

\subsection{Notation and Evaluation Horizons}

Table~\ref{tab:method_notation} summarizes the notation used throughout.

\begin{table}[tbp]
\centering
\footnotesize
\renewcommand{\arraystretch}{1.05}
\begin{tabular}{@{}p{0.25\columnwidth} p{0.67\columnwidth}@{}}
\hline
Symbol & Definition \\
\hline
$i$ & enrollment (unit of analysis) \\
$t$ & discrete week \\
$X_{it}$ & observed covariates at time $t$ \\
$E_i$ & primary event indicator \\
$C_i$ & censoring time \\
$\hat h_{it}$ & predicted weekly hazard \\
$\hat S_i(t)$ & predicted survival \\
$\bar S^{(a)}(t)$ & mean survival under scenario $a$ \\
$\Delta S(t)$ & structural survival contrast \\
$Gap^{(a)}(t)$ & between-group gap under scenario $a$ \\
$\Delta Gap(t)$ & change in gap between scenarios \\
$T_{policy}$ & primary substantive reporting horizon \\
$T_{eval\_metrics}$ & stable horizon for IPCW-based metrics \\
$g_{min}$ & minimum stability threshold for $\hat G(t)$ \\
\hline
\end{tabular}
\caption{Main notation. For the reported OULAD run: $T_{policy}=18$, $T_{eval\_policy}=38$, $T_{eval\_metrics}=37$, $g_{min}=0.05$.}
\label{tab:method_notation}
\end{table}

\subsection{Problem Setup, Endpoint, and Person-Period Construction}

The unit of analysis is enrollment $i$ observed over discrete weekly time $t$. The primary event is administrative withdrawal:
\begin{equation*}
E_i=\mathbb{1}\{\texttt{final\_result}_i=\texttt{Withdrawn}\ \land\ \texttt{date\_unregistration}_i\ \text{is valid}\}.
\end{equation*}
Each enrollment is expanded into weekly rows $t=0,\dots,t_i^{\mathrm{final}}$ with terminal event label $event_{it}=\mathbb{1}\{E_i=1 \land t=t_i^{\mathrm{final}}\}$, yielding a longitudinal risk set with at most one terminal event week per enrollment.

\paragraph*{Temporal covariate engineering.}
Dynamic covariates are defined on a weekly grid under temporally safe logic. Key features include \texttt{total\_clicks} (operational engagement proxy), recency counter (consecutive weeks since last activity), and activity streak. These encode short-term engagement dynamics while preserving temporal ordering and preventing leakage.

\subsection{Primary Discrete-Time Hazard Model (RQ1)}

Following standard discrete-time event-history formulations~\cite{SingerWillett1993DiscreteTimeSurvival,Allison1982DiscreteTimeMethods}:
\begin{equation}
\label{eq:hazard_weekly}
h_{it}=P(event_{it}=1 \mid T_i\ge t, X_{it}), \quad \text{logit}(\hat h_{it})=\beta_0+\beta^\top X_{it}
\end{equation}
\begin{equation}
\label{eq:survival_product}
\hat S_i(t)=\prod_{k\le t}(1-\hat h_{ik}).
\end{equation}
Raw weekly probabilities are post-hoc calibrated through Platt scaling~\cite{Platt1999ProbabilisticOutputs} under enrollment-respecting split logic, yielding interpretable hazard probabilities over time.

\paragraph*{Product-form rank compression.}
\begin{proposition}[Product-form rank compression]
\label{prop:rank_inversion}
Let enrollment $i$ be observed over weeks $t\in\mathcal{T}_i\subseteq\{0,\dots,T\}$ with hazards $\hat h_{it}\in[0,1)$ and product-form survival $\hat S_i(T)=\prod_{t\le T}(1-\hat h_{it})$; let its horizon risk score be $r_i=1-\hat S_i(T)$. Then $r_i$ is a strictly increasing function of the cumulative hazard $H_i(T)=\sum_{t\le T}\hat h_{it}$. Because $H_i(T)$ is a sum over observed weeks, it conflates the per-week risk level with the number of observed weeks $|\mathcal{T}_i|$: for equal mean weekly hazard, an enrollment observed over more weeks attains a larger $H_i(T)$, hence a larger $r_i$. Consequently, whenever observation length is not aligned with horizon event status, the ranking induced by $r_i$ may invert relative to the ranking by mean weekly hazard $\bar h_i=|\mathcal{T}_i|^{-1}H_i(T)$.
\end{proposition}

\subsection{Policy Simulation (RQ2)}
\label{subsec:policy_sim}

The operational intervention rule is: ``flag if no LMS engagement in the last 7~days ($\text{recency} \ge 1$ week)'', with trigger parameter $r^*=1$ and active window $W=2$~weeks, anchored to the nudge framing of \citeA{KayBostock2023PowerNudge}:
\begin{equation*}
t_i^*=\min\{t: Recency_{it}\ge r^*\}, \quad active_{it}=\mathbb{1}\{t\in[t_i^*,t_i^*+W)\}.
\end{equation*}
Under the \textit{shock} scenario (sensitivity/stress-test device):
\begin{equation*}
\hat h^{(1,shock)}_{it}=\begin{cases}\hat h^{(0)}_{it}(1-\delta_{shock}), & active_{it}=1, \\ \hat h^{(0)}_{it}, & active_{it}=0.\end{cases}
\end{equation*}
Under the \textit{mechanism-aware} scenario (substantive estimand), the policy propagates through hypothetical covariate updates: $\hat h^{(1,mech)}_{it}=f(X^{cf}_{it})$, where active-window rows are updated by setting recency to~$0$ and flooring streak at~$1$~\cite{KayBostock2023PowerNudge,Ahmadi2023LMSEngagementIndicators,Nkomo2021StudentEngagementDigitalTech}, modelling one week of engagement in response to a nudge rather than assuming sustained multi-week recovery.

Effect intensity values are analyst-chosen scenario parameters, not estimated effect sizes (\S\ref{para:interp_contract}).

\subsection{Structural Survival Contrast and Fairness Track (RQ2--RQ3)}

\begin{equation}
\label{eq:delta_s}
\Delta S(t)=\bar S^{(1)}(t)-\bar S^{(0)}(t), \quad \bar S^{(a)}(t)=\frac{1}{N}\sum_i \hat S_i^{(a)}(t).
\end{equation}
The policy layer is a deterministic operator $\mathcal{O}:(\hat h_{it},\theta_{\text{policy}})\mapsto \Delta S(t)$---an exact transformation of model outputs (\S\ref{para:interp_contract}).

For subgroup analysis (RQ3):
\begin{equation}
\label{eq:delta_gap}
\Delta Gap(t)=Gap^{(1)}(t)-Gap^{(0)}(t), \quad Gap^{(a)}(t)=\bar\mu_1^{(a)}(t)-\bar\mu_0^{(a)}(t).
\end{equation}
Uncertainty is estimated via $B=500$ enrollment-level bootstrap resamples~\cite{PessachShmueli2022FairnessReview,Lu2022ReliabilityFairnessAudit}.

\subsection{Censoring and IPCW}

IPCW reweights observations to correct for potentially informative censoring:
\begin{equation*}
w_{it}=\frac{1}{\max(\hat G_{it}, g_{min})}, \quad \hat G_{it}=P(C_i\ge t \mid X_{it}).
\end{equation*}
This follows standard survival-prediction evaluation under right censoring~\cite{Graf1999AssessmentSurvivalPrediction,Gerds2006ConsistentBrier}. Because non-event follow-up is anchored to the last observed VLE week, the conditional-independence assumption required for IPCW consistency is likely violated; the correction is at best approximate, and IPCW-weighted metrics are best interpreted as descriptive sensitivity bounds.

\subsection{Reproducibility}

Code for all pipeline stages is available at~\cite{RafaRepoTCMStudentDropout}. The reference dataset is OULAD~\cite{Kuzilek2017OULAD}. Modeling choices follow the spirit of model card documentation practices~\cite{Mitchell2019ModelCards}.

\section{Experimental Design}
\label{sec:exp_design}

\subsection{Cohort and Person-Period Table}
\label{subsec:cohort}

We use OULAD~\cite{Kuzilek2017OULAD}, integrating time-stamped VLE interaction logs (\texttt{studentVle}), assessment records, and administrative enrollment data (\texttt{studentInfo}, \texttt{studentRegistration}). After deduplication, the exported cohort contains $N=32{,}593$ enrollments from $28{,}785$ unique students spanning 7 course presentations across 2013--2014 academic years.

\begin{table}[tbp]
\centering
\caption{Cohort summary and train/test partition after stratified temporal split.}
\label{tab:cohort_split}
\begin{tabular}{lrr}
\hline
Quantity & Train & Test \\
\hline
Enrollments & 22{,}815 & 9{,}778 \\
Events (Withdrawn) & 5{,}171 & 2{,}216 \\
Person-period rows & 542{,}878 & 232{,}417 \\
\hline
\end{tabular}
\end{table}

\subsection{Stratified Temporal Split}
\label{subsec:split}

The train-test split is performed at the enrollment level to prevent row leakage. Stratification uses event status and 4-quantile temporal bucket (bucket edges: $[0,8,31,34,63]$). Within each stratum, 30\% of enrollments are assigned to the test set (seed~42). This design preserves both event-status balance and coarse temporal structure across partitions, consistent with cross-validation best practices for dependent data~\cite{KapoorNarayanan2023LeakagePatterns,RobertsEtAl2017CrossValidationEcography}. The test set contains 2{,}216 events, of which 1{,}637 occur by $T_{policy}=18$ (74\%), exceeding the commonly cited minimum of 10 events per predictor~\cite{Peduzzi1996EPVLogistic}.

\subsection{Model Variants and Evaluation Metrics}
\label{subsec:model_variants}

The primary model is penalized logistic regression with Platt calibration (\texttt{solver=liblinear}, \texttt{C=1.0}, \texttt{class\_weight=balanced}, GroupKFold $k=5$, enrollment-grouped)~\cite{Kull2017BetaCalibration}. Benchmark models (enrollment-level static LR, RSF~\cite{MogensenIshwaranGerds2012RSF}, DeepHit~\cite{LeeDeepHit2018}) use static enrollment-level features without weekly trajectory inputs, isolating the contribution of temporal structure. All benchmarks use library-default hyperparameters without task-specific tuning.

\paragraph*{Recalibrated mean-hazard variant.}
The recalibrated mean-hazard variant is an enrollment-level prediction derived from the primary model without retraining: $\bar h_i = |\mathcal{T}_i|^{-1}\sum_{t \in \mathcal{T}_i} \hat h_{it}$, re-calibrated to the event-at-horizon label via Platt scaling. By avoiding the product-form survival collapse identified in Proposition~\ref{prop:rank_inversion}, it recovers reliable enrollment-level ranking and attains $\text{AUC}_{IPCW}(T_{policy})=0.7748$.

Evaluation horizons: $T_{policy}=18$ (primary), $T_{eval\_metrics}=37$ (stable IPCW; last week with $\hat G\ge 0.05$), $T_{eval\_policy}=38$ (trajectory visualization only). Primary metrics: $AUC_{row}$, IPCW Brier, IBS~\cite{Graf1999AssessmentSurvivalPrediction}, ECE$_{15}$~\cite{Naeini2015ECECalibration}, C-index~\cite{Uno2011CStatisticsCensored}.

\subsection{Policy Evaluation Protocol (RQ2)}
\label{subsec:policy_eval}

The scenario catalog spans 19 shock intensities ($\delta_{shock}\in[0.02,0.75]$; anchored conservative $\delta_{shock}=0.08$, which falls within the medium-effect band for education interventions~\cite{Kraft2020EffectSizesEducation,EvansYuan2022EffectSizesInternational}) and a sensitivity grid of 216 trigger/schedule combinations. A shared mechanism-aware schedule ($\alpha_{week0}=0.35$, $\alpha_{week1}=0.10$, \texttt{kb2023\_step\_2w} decay~\cite{KayBostock2023PowerNudge}) is fixed across all scenarios. Bootstrap for $\Delta S$: $B=500$ enrollment-level resamples, seed~42, percentile 95\% CI.

\subsection{Subgroup Evaluation Protocol (RQ3)}
\label{subsec:fairness_eval}

Subgroups analyzed: gender (M/F; $g=0$=Male, $g=1$=Female) and disability status (N/Y). No subgroup-specific preprocessing or calibration is applied: the same fitted hazard model is evaluated identically across all groups. Bootstrap: $B=500$ enrollment-level resamples, seed~42, percentile 95\% CI. A fairness signal is considered robust when $0\notin CI_{95\%}(\Delta Gap(T))$.

\section{Empirical Evaluation: Hazard Quality, Policy Contrasts, and Subgroup Analysis}

\subsection{Direct Answers to the Research Questions}
\label{subsec:direct_answers}

\begin{table}[tbp]
\centering
\scriptsize
\renewcommand{\arraystretch}{1.05}
\caption{Direct answers to RQ1--RQ3. Group coding: F=female, M=male; Y=disability declared, N=no declaration. Bootstrap: $B=500$, percentile 95\% CI, seed~42. Horizons: $T_{policy}=18$; $T_{eval}\equiv T_{eval\_metrics}=37$.}
\label{tab:rq_answers}
\begin{tabular}{@{}>{\raggedright\arraybackslash}p{0.09\columnwidth}
                >{\raggedright\arraybackslash}p{0.86\columnwidth}@{}}
\hline
RQ & Direct answer \\
\hline
RQ1 & Yes, for temporal risk ranking, with caution in the highest-risk tail. $AUC_{row,test}=0.8396$; $\text{ECE}_{15}=0.0012$; recalibrated mean-hazard $\text{AUC}_{IPCW}(T_{policy})=0.7748$. \\
RQ2 & The operator returns a monotone, sign-stable dose--response; a positive shock contrast is definitional given a hazard-reducing scenario, so the reportable content is the auditable response, not the sign. Structural survival contrasts at $T_{policy}=18$: anchored ($\delta=0.08$) $\Delta S=+0.0108$; hypothetical~A ($\delta=0.20$) $+0.0274$; hypothetical~B ($\delta=0.60$) $+0.0865$; mech-aware $+0.0121$. \\
RQ3 & A capability demonstration, not an equity finding: the monitor returns directionally stable but negligible gap changes ($|\Delta Gap|\approx 5\times 10^{-4}$). Gender (F$-$M): $\Delta Gap(18)=-0.000508$ CI$(-0.000662,-0.000329)$. Disability (Y$-$N): $\Delta Gap(18)=-0.000402$ CI$(-0.000661,-0.000106)$. All four bootstrap CIs exclude zero; detectability at this magnitude does not imply an actionable equity change. \\
\hline
\end{tabular}
\end{table}

\subsection{RQ1: Temporal Hazard Quality and Actionable Timing}
\label{subsec:rq1_results}

Row-level weekly hazard discrimination reaches $AUC_{test}=0.8396$ under the stratified temporal holdout, with aggregate calibration $\text{ECE}_{15}=0.0012$. Both pre-specified operational criteria ($AUC_{row,test}\ge 0.80$, $\text{ECE}_{15}\le 0.01$) are met.

\paragraph*{Discrimination.}
Per-week diagnostics confirm stability across 35 event-populated weeks ($AUC_{test}>AUC_{train}$ in 22 of 35 weeks, 63\%). At $T_{policy}=18$, 1{,}637 of 2{,}216 test events (74\%) are observed.

\paragraph*{Which AUC to trust.}
The row-level $AUC_{row}=0.8396$ is computed over person-period rows, the large majority of which are low-hazard ``still enrolled'' weeks; it is the appropriate metric for the RQ1 weekly-ranking criterion, but it flatters the model relative to enrollment-level decision-making. The honest figure for ranking \emph{enrollments} at a horizon is the IPCW horizon AUC, $\text{AUC}_{IPCW}(T_{policy})=0.7748$, and even that value is attained only after the mean-hazard recalibration that corrects the product-form rank inversion the temporal survival representation itself induces (\S\ref{subsec:benchmark_robust_results}; Proposition~\ref{prop:rank_inversion}). Against the static enrollment-level baselines (non-temporal RSF $0.663$, static LR $0.659$; Table~\ref{tab:benchmark_horizon_results}), this recalibrated temporal model buys roughly $+0.11$ AUC---a real but modest gain, and one the un-recalibrated temporal representation does not deliver on its own. Enrollment-level claims in this paper should therefore be weighted on the $0.77$ figure, not the $0.84$ one.

\begin{figure}[tbp]
\centering
\includegraphics[width=\columnwidth]{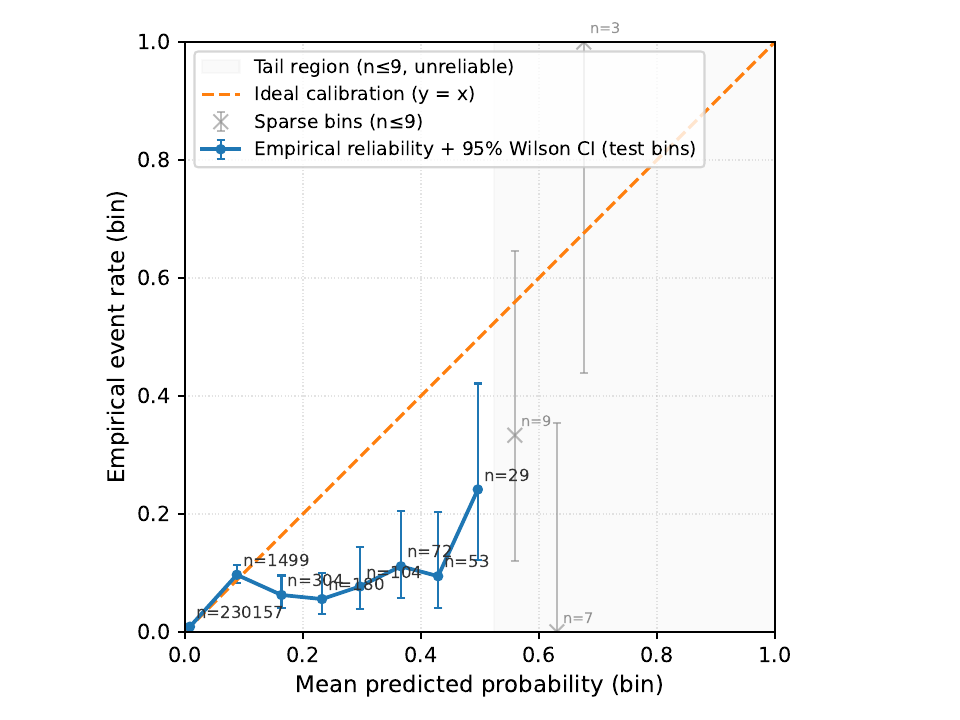}
\caption{Weekly hazard calibration on the test set (15 equal-width bins; 95\% Wilson CI per bin; bin counts annotated). Bins~8--10 contain $\le 9$ rows each (Table~\ref{tab:calibration_tail_support}). Aggregate calibration governed by Bin~0 ($n=230{,}157$; $\text{ECE}_{15}=0.0012$).}
\label{fig:calibration_main}
\end{figure}

\paragraph*{Calibration.}
Figure~\ref{fig:calibration_main} provides the primary calibration evidence for RQ1.

Aggregate calibration is tight ($\text{ECE}_{15}=0.0012$) with sparse support in the highest-risk tail (Bins~8--10, combined $n=19$; Table~\ref{tab:calibration_tail_support}); calibration estimates above $\hat h_{it}\ge 0.53$ are unreliable and should not be used for high-risk student targeting without additional external validation.

\begin{table}[tbp]
\centering
\caption{Support of highest-risk calibration bins on test rows.}
\label{tab:calibration_tail_support}
\begin{tabular}{lcccc}
\hline
Bin & Risk interval & $n$ & Events & Non-events \\
\hline
Bin 8 & [0.533, 0.600) & 9 & 3 & 6 \\
Bin 9 & [0.600, 0.667) & 7 & 0 & 7 \\
Bin 10 & [0.667, 0.733) & 3 & 3 & 0 \\
\hline
\end{tabular}
\end{table}

\begin{figure}[tbp]
\centering
\includegraphics[width=0.82\columnwidth]{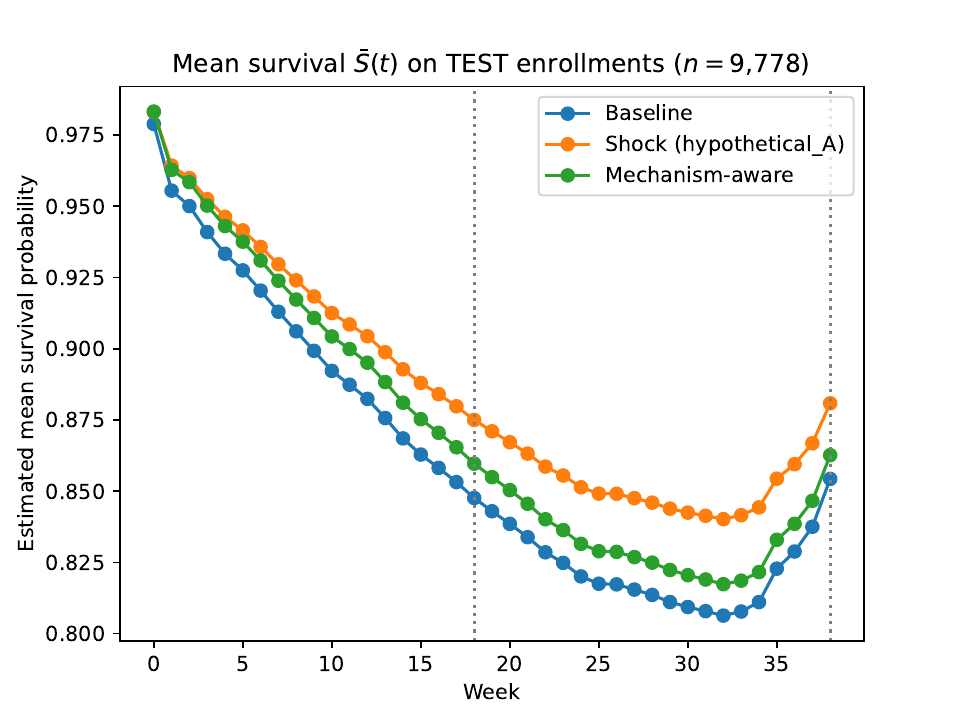}
\caption{Mean survival trajectories on the test set ($n=9{,}778$ enrollments). Lines show baseline, shock (hypothetical~A, $\delta_{shock}=0.20$, shown for visual clarity), and mechanism-aware scenario means. Vertical dotted lines mark $T_{policy}=18$ and $T_{eval\_policy}=38$. Primary reported scenario is the anchored conservative ($\delta_{shock}=0.08$, reported numerically in Table~\ref{tab:rq2_summary}).}
\label{fig:survival_baseline_rq1}
\end{figure}

\paragraph*{Temporal decision window.}
Figure~\ref{fig:survival_baseline_rq1} shows the baseline mean survival trajectory. The implied mean hazard peaks at $t=1$ ($\bar h(1) \approx 0.024$), with a secondary cluster at weeks~3--4 ($\bar h \approx 0.009$--$0.010$); the highest-risk window is concentrated in the first four weeks.

\paragraph*{RQ1 limitation.}
Sparse support in the highest-risk bins ($\hat h_{it}\ge 0.20$; $n\le 9$ rows; Table~\ref{tab:calibration_tail_support}) means calibration in that region cannot be reliably estimated. This model should \emph{not} be used for targeted intervention based solely on high-risk predicted scores without additional validation in this range.

\subsection{Censoring Validity and the Dual-Horizon Rule}
\label{subsec:censoring_results}

\begin{figure}[tbp]
\centering
\includegraphics[width=0.82\columnwidth]{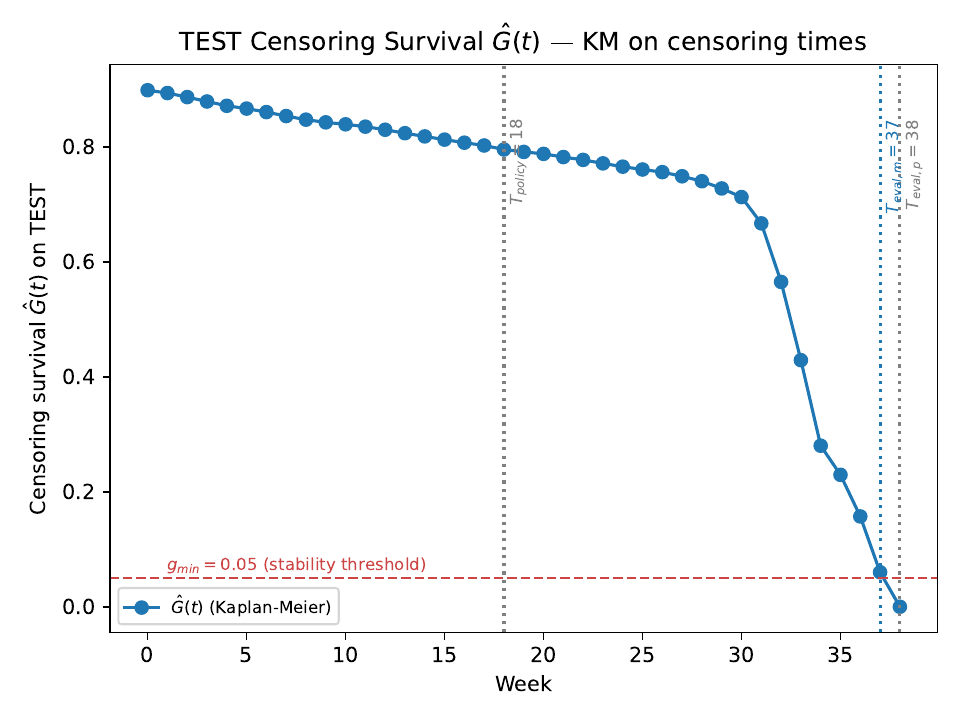}
\caption{Censoring survival $\hat G(t)$ on the test set ($N=9{,}778$ enrollments). Vertical lines mark the three evaluation horizons; horizontal line at $g_{min}=0.05$ marks the IPCW stability cutoff.}
\label{fig:censoring_survival_results}
\end{figure}

Figure~\ref{fig:censoring_survival_results} displays the estimated censoring survival $\hat G(t)$ on the test set. Weighted metrics are interpreted only where censoring support holds: $\hat G(18)=0.796$ (strong; primary horizon), $\hat G(37)=0.060$ (near threshold; IPCW-metric horizon), $\hat G(38)\approx 10^{-12}$ (unstable; trajectory visualization only).

\subsection{RQ2: Policy Simulation and Structural Scenario Contrast}
\label{subsec:rq2_results}

\paragraph*{What RQ2 does and does not establish.}
The sign of the shock contrast is not an empirical result: because product-form survival is monotone in each weekly hazard, any hazard-reducing scenario mechanically yields $\Delta S>0$, and its magnitude scales with the assumed intensity $\delta_{shock}$, which is fixed by analyst choice rather than estimated. What RQ2 establishes is therefore (i)~that the operator produces a \emph{monotone, inspectable} dose--response under an explicit, re-runnable trigger/schedule contract; (ii)~that the substantive mechanism-aware estimand---which propagates the intervention through the fitted covariate path rather than overwriting the model output---is sign-stable across the robustness grid; and (iii)~that both can be reported at a censoring-valid horizon with bootstrap uncertainty. The numerical values below are the output of that operator, not evidence that the modeled policy would reduce real dropout.

\begin{table}[tbp]
\centering
\caption{RQ2 summary by scenario at $T_{policy}=18$ and $T_{eval\_policy}=38$ (point estimates). Bootstrap 95\% CIs ($B{=}500$, seed~42, percentile) exported to \nolinkurl{rq2_policy_bootstrap_ci.csv}.}
\label{tab:rq2_summary}
\small
\begin{tabular}{lccccc}
\hline
Scenario & $\delta_{shock}$ & $\Delta S_{shock}(18)$ & $\Delta S_{shock}(38)$ & $\Delta S_{mech}(18)$ & $\Delta S_{mech}(38)$ \\
\hline
Anchored conservative & 0.08 & 0.0108 & 0.0105 & +0.0121 & +0.0083 \\
Hypothetical A        & 0.20 & 0.0274 & 0.0266 & +0.0121 & +0.0083 \\
Hypothetical B        & 0.60 & 0.0865 & 0.0834 & +0.0121 & +0.0083 \\
\hline
\end{tabular}
\end{table}

Translating to institutional scale: the anchored $\Delta S(18)=+0.0108$ corresponds to approximately 106 additional enrollments projected to survive to week~18 across the $9{,}778$ test enrollments---a model-implied projection conditional on the assumed $\delta_{shock}$ (\S\ref{para:interp_contract}). The 216-point sensitivity grid confirms directional consistency across scenario families.

\subsection{Benchmark and Robustness Evidence}
\label{subsec:benchmark_robust_results}

\begin{table}[tbp]
\centering
\caption{Cross-family IPCW benchmark at reported horizons. $^\dag$Product-form survival rank inversion (Proposition~\ref{prop:rank_inversion}); AUC$<0.5$ expected. Values are point estimates from a single stratified temporal holdout.}
\label{tab:benchmark_horizon_results}
\small
\begin{tabular}{lcc}
\hline
Model family & AUC$_{IPCW}(T_{policy})$ & AUC$_{IPCW}(T_{eval\_metrics})$ \\
\hline
Temporal hazard$^\dag$                  & 0.4768 & 0.4950 \\
Temporal hazard + IPCW$^\dag$           & 0.5220 & 0.5291 \\
Enrollment-level LR (static)            & 0.6586 & 0.6479 \\
Non-temporal RSF                        & 0.6634 & 0.6520 \\
DeepHit (static enrollment only)        & 0.6228 & 0.5947 \\
\textbf{Temporal hazard (recalibrated)} & \textbf{0.7748} & \textbf{0.7714} \\
\hline
\end{tabular}
\end{table}

The recalibrated temporal hazard (mean-hazard aggregation per Proposition~\ref{prop:rank_inversion}) achieves AUC$_{IPCW}(T_{policy})=0.7748$, recovering the enrollment-level discrimination lost through rank inversion ($\Delta\text{AUC}\approx+0.30$ relative to the un-recalibrated temporal model). A static logistic regression achieves AUC$_{event}(T_{policy})=0.659$ and the non-temporal RSF reaches $0.663$. The static-only DeepHit baseline (0.623) confirms that a neural architecture without temporal-trajectory inputs yields no enrollment-level discrimination advantage here. Net, the temporal representation's enrollment-level edge over the best static baseline is approximately $+0.11$ AUC ($0.7748$ vs.\ $0.663$) and is realized only after mean-hazard recalibration; before that correction the static baselines rank enrollments better than the temporal model.

\paragraph*{Ablation.}
Removing Recency/Streak causes the largest AUC drop ($AUC_{row}=0.7987$ vs.\ $0.8396$ for the full model); removing \texttt{total\_clicks} causes a smaller reduction ($0.8263$). Five held-out course runs confirm $\text{AUC}_{row}>0.76$ across all five (range: 0.765--0.839).

\subsection{RQ3: Subgroup Validation via Change-in-Gap}
\label{subsec:rq3_results}

RQ3 evaluates the framework's \textit{capability} to produce subgroup-sensitive structural contrasts. The reported magnitudes are not offered as substantive equity evidence but as a demonstration that the contrast tooling is operational.

\begin{table}[tbp]
\centering
\caption{RQ3 summary: $\Delta Gap(T)$ with bootstrap uncertainty ($B=500$, seed~42, percentile 95\% CI). Group1$-$Group0: gender Group0=M, Group1=F; disability Group0=N, Group1=Y.}
\label{tab:rq3_summary}
\small
\begin{tabular}{llcc}
\hline
Attribute & Horizon & $\Delta Gap(T)$ & 95\% bootstrap CI \\
\hline
Gender (F$-$M)      & $T_{policy}=18$        & $-0.000508$ & $(-0.000662,\,-0.000329)$ \\
                    & $T_{eval\_metrics}=37$ & $-0.000567$ & $(-0.000715,\,-0.000407)$ \\
Disability (Y$-$N)  & $T_{policy}=18$        & $-0.000402$ & $(-0.000661,\,-0.000106)$ \\
                    & $T_{eval\_metrics}=37$ & $-0.000446$ & $(-0.000687,\,-0.000194)$ \\
\hline
\end{tabular}
\end{table}

All four bootstrap intervals exclude zero at both reported horizons (Table~\ref{tab:rq3_summary}), with $P(\Delta Gap<0)=100\%$ for gender and $\ge 99.4\%$ for disability. The $|\Delta Gap|\approx 5\times 10^{-4}$ should be interpreted relative to the absolute baseline survival gap between groups: statistical detectability does not necessarily translate to actionable fairness improvement.

\paragraph*{Equity note.}
The Recency trigger ($r^*\ge 1$ week of inactivity) may disproportionately activate for students whose engagement patterns are constrained by external factors (part-time employment, caregiving, constrained technology access) rather than by disengagement per se~\cite{Summers2022DisadvantageEngagement}. A more comprehensive evaluation could extend to calibration parity, equalized-odds audits over time, and additional sensitive attributes.

\subsection{Results Synthesis}
\label{subsec:results_synthesis}

Together, the three result blocks form a coherent evidence chain. Weekly hazard quality (RQ1) establishes the ranking foundation that makes scenario contrasts ($\Delta S(t)$, RQ2) meaningful; the same survival model supports subgroup-sensitive gap estimation ($\Delta Gap(T)$, RQ3). The main interpretation remains directionally coherent under benchmark and robustness axes.

The reported results span two evidential classes: (a)~empirical observations supported directly by OULAD test data (weekly hazard discrimination, row-level calibration, censoring-survival curve); and (b)~model-conditional structural contrasts ($\Delta S(t)$, $\Delta Gap(T)$) that carry the uncertainty of the generative model in addition to sampling uncertainty.

The enrollment-level rank inversion (AUC$_{IPCW}<0.5$; Table~\ref{tab:benchmark_horizon_results}) is the product-form rank compression of Proposition~\ref{prop:rank_inversion}. The recalibrated mean-hazard variant normalises by the number of observed weeks, removing the length term, and recovers AUC$_{IPCW}(T_{policy})=0.7748$.

\begin{enumerate}[noitemsep]
  \item Weekly hazard discrimination ($AUC_{test}=0.8396$) and aggregate calibration ($\text{ECE}_{15}=0.0012$) support temporal risk ranking and trajectory analysis, with caution in the highest-risk tail bins (RQ1).
  \item Structural survival contrasts $\Delta S(t)$ are positive under both scenario branches at the reported horizons (Table~\ref{tab:rq2_summary}; RQ2).
  \item The same policy can be audited for subgroup differentials via $\Delta Gap(T)$; the reported contrasts are directionally stable but negligible ($|\Delta Gap| \approx 5 \times 10^{-4}$), establishing that the monitor is operational rather than that the policy changes equity (RQ3).
  \item The main interpretation remains directionally coherent under benchmark and robustness axes.
\end{enumerate}

\subsection{Interpretation Boundaries}

Three boundaries govern every reading of the results: (i)~policy and fairness outputs are model-implied structural contrasts, not causally identified effects (\S\ref{para:interp_contract}); (ii)~censoring support is binding for metric interpretation, so weighted metrics are read only where $\hat G(t)\ge g_{min}$; and (iii)~endpoint and domain-shift checks indicate sensitivity ranges, not universal transportability. Within these limits, a reproducible temporal pipeline translates weekly risk into structured policy and subgroup diagnostics.

\section{Conclusion}

This paper introduced a reproducible, auditable framework that translates estimated weekly risk into structural contrasts of intervention policies and subgroup-equity diagnostics on intervention-free observational data. The three result blocks form a chain: weekly hazard quality (RQ1) establishes the ranking foundation; scenario contrasts ($\Delta S(t)$, RQ2) quantify structural comparisons; and subgroup-sensitive gap estimation ($\Delta Gap(T)$, RQ3) provides the equity diagnostic. The recalibrated mean-hazard variant resolves the enrollment-level rank inversion, recovering $\text{AUC}_{IPCW}(T_{policy})=0.7748$ and making RQ2 horizon comparisons tractable.

\paragraph*{RQ answers.}
\textbf{RQ1:} $AUC_{row,test}=0.8396$ and $\text{ECE}_{15}=0.0012$ meet both pre-specified criteria; enrollment-level horizon discrimination is recovered to $\text{AUC}_{IPCW}=0.7748$ by the recalibrated mean-hazard variant. \textbf{RQ2:} Positive structural survival contrasts under both scenario branches at both reported horizons ($\Delta S_{mech}(18)=+0.0121$; anchored $\Delta S_{shock}(18)=+0.0108$), with directional consistency across the 216-point robustness grid. \textbf{RQ3:} Differential subgroup gap changes detected for gender and disability ($|\Delta Gap|\approx 5\times 10^{-4}$, all four bootstrap CIs excluding zero); magnitude is negligible and should be interpreted as operational feasibility evidence, not substantive equity improvement.

\subsection{Limitations}

Four limitations bound the reported results. (1)~\textit{Non-causal interpretation}: the observational design does not permit identification of unique intervention effect sizes; $\Delta S(t)$ and $\Delta Gap(T)$ are structural model outputs, not estimates of real treatment effects. (2)~\textit{Informative censoring}: because non-event follow-up is anchored to the last observed VLE week, the CIA assumption required for IPCW consistency is likely violated; the tipping-point analysis ($\kappa^*>1.27$ at $T_{policy}$) provides robustness evidence but the IPCW correction is approximate. (3)~\textit{Evaluation scope}: the discrete-time logistic framework with five-run within-OULAD generalization test bounds the evaluation; results should not be extrapolated to sequence-aware models or datasets outside the OULAD distribution without re-evaluation. (4)~\textit{Single institution}: OULAD represents The Open University, UK (predominantly distance-learning, non-traditional population); the reported results should be tested on at least one additional dataset before broader deployment claims can be made.

\subsection{Practical Implications}

The framework supports three institutional use cases: (i)~\emph{weekly risk dashboards}: hazard scores $\hat h_{it}$ surfaced to advisors as a ranked watchlist; (ii)~\emph{intervention trigger protocols}: the Recency trigger ($\text{Recency}\ge r^*=1$) with a two-week active window defines an auditable, inspectable rule that can be logged and re-evaluated without model retraining; (iii)~\emph{equity monitoring}: the $\Delta Gap(T)$ diagnostic provides a lightweight audit of whether a trigger rule changes survival gaps across demographic groups, with bootstrap uncertainty.

Future work could extend the temporal backbone to sequence-aware recurrent variants, evaluated under the same stratified holdout and IPCW protocol against the $\text{AUC}_{IPCW}(T_{policy})=0.7748$ baseline. The policy simulation layer could also be expanded to multi-group fairness-constrained optimization, formulating scenario selection as a constrained program that maximizes $\Delta S(T_{policy})$ subject to $|\Delta Gap(T)|\le\epsilon$.

\appendix

\section{Appendix A: Diagnostics and Robustness Audits}
\label{app:a}

\subsection{A.1 Bootstrap Uncertainty for RQ2}
\label{app:a_rq2_bootstrap}

\begin{table}[tbp]
\centering
\caption{Bootstrap uncertainty for RQ2 at $T_{policy}=18$ ($B=500$, seed~42, percentile). $^\dagger$Bootstrap means differ from Table~\ref{tab:rq2_summary} point estimates. Full 19-scenario table: \nolinkurl{rq2_policy_bootstrap_ci.csv}.}
\label{tab:app_boot_rq2}
\small
\begin{tabular}{lccc}
\hline
Scenario & $\Delta S^\dagger_{shock}(18)$ mean & 95\% CI$_{lo}$ & 95\% CI$_{hi}$ \\
\hline
Anchored conservative & 0.00974 & 0.00962 & 0.00986 \\
Hypothetical A        & 0.02467 & 0.02437 & 0.02498 \\
Hypothetical B        & 0.07753 & 0.07650 & 0.07854 \\
\multicolumn{4}{l}{Mechanism-aware (all scenarios): 0.01473 [0.01446, 0.01501]} \\
\hline
\end{tabular}
\end{table}

\subsection{A.2 Bootstrap Uncertainty for RQ3}
\label{app:a_rq3_bootstrap}

\begin{table}[tbp]
\centering
\caption{Bootstrap distribution summary of $\Delta Gap(T)$ ($B=500$, seed~42). $P(\Delta Gap{<}0)$: fraction of replicates with negative gap reduction.}
\label{tab:app_boot_tp}
\small
\begin{tabular}{llccc}
\hline
Attribute & Horizon & $\widehat{\Delta Gap}$ & 95\% bootstrap CI & $P(\Delta Gap{<}0)$ \\
\hline
Gender (F$-$M) & $T_{policy}=18$ & $-0.000505$ & $(-0.000662,\,-0.000329)$ & 100.0\% \\
               & $T_{eval}=37$   & $-0.000566$ & $(-0.000715,\,-0.000407)$ & 100.0\% \\
Disability (Y$-$N) & $T_{policy}=18$ & $-0.000391$ & $(-0.000661,\,-0.000106)$ & 99.4\% \\
                   & $T_{eval}=37$   & $-0.000441$ & $(-0.000687,\,-0.000194)$ & 99.8\% \\
\hline
\end{tabular}
\end{table}

\subsection{A.3 By-Group Predictive Diagnostics}

\begin{table}[tbp]
\centering
\caption{By-group row-level diagnostics. Gender (top); disability (bottom).}
\label{tab:app_group_metrics}
\begin{tabular}{lccc}
\hline
Group & AUC$_{row}$ & Brier & ECE \\
\hline
Female (F)        & 0.8391 & 0.0089 & 0.0018 \\
Male (M)          & 0.8407 & 0.0098 & 0.0020 \\
\hline
No disability (N) & 0.8401 & 0.0090 & 0.0016 \\
Disability (Y)    & 0.8376 & 0.0136 & 0.0059 \\
\hline
\end{tabular}
\end{table}

\subsection{A.4 Censoring Anchor Sensitivity}
\label{app:a_censoring}

\begin{table}[tbp]
\centering
\small
\caption{Censoring anchor sensitivity: censoring model test AUC and capped-weight share.}
\label{tab:censoring_anchor_sensitivity}
\begin{tabular}{lcc}
\toprule
Anchor variant & Cens.~AUC (test) & Capped-weight share \\
\midrule
Conservative (\texttt{anchor\_-1}) & 0.9163 & 0.5029 \\
Default (\texttt{anchor\_0})       & 0.9299 & 0.6576 \\
\bottomrule
\end{tabular}
\end{table}

The high censoring-model AUC (0.92--0.93) confirms informative censoring. IPCW-weighted metrics at $T_{eval\_metrics}=37$ are best treated as conservative lower bounds on true calibration error rather than definitive estimates.

\section{Appendix B: Executable Pipeline and Robustness Battery}
\label{app:b}

\subsection{B.1 Planned Robustness Battery}

Three ablation variants (Full; No Recency/Streak; No Activity), five held-out course runs (AUC$_{row}$ range: 0.765--0.839), and an endpoint sensitivity check (primary Withdrawn vs.\ composite Fail$\lor$Withdrawn) form the robustness battery pre-specified in \S\ref{subsec:fairness_eval}.

\subsection{B.2 End-to-End Pipeline}
\label{app:b_pipeline}

\begin{algorithm}[!t]
\caption{Discrete-Time Survival + Scenario-Based Policy Simulation}
\label{alg:pipeline}
\small
\begin{algorithmic}[1]
\Require OULAD: \texttt{studentInfo}, \texttt{studentRegistration}, \texttt{studentVle}
\Ensure $\hat S^{(0)}(t)$, $\hat S^{(1)}(t)$, $\Delta S(T_{policy})$, $\Delta Gap(T)$ with bootstrap CI
\State Build enrollment backbone keyed by $(\texttt{id\_student},\texttt{code\_module},\texttt{code\_presentation})$.
\State Define endpoint: $E_i=\mathbb{1}[\texttt{Withdrawn} \wedge \texttt{date\_unregistration valid}]$.
\State Compute censoring anchor $t^{last}_i$ (last observed VLE week).
\State Construct weekly person-period rows $t=0,\dots,t^{final}_i$.
\State Engineer weekly covariates (clicks, Recency, Streak, submitted).
\State Perform enrollment-stratified temporal split ($q=4$, \texttt{test\_size}=0.30, seed~42).
\State Fit discrete-time hazard model on training rows (penalized logistic + Platt calibration).
\State Compute $\hat h^{(0)}_{it}$ on test rows; reconstruct $\hat S^{(0)}_i(t)=\prod_{k\le t}(1-\hat h^{(0)}_{ik})$.
\State Apply Recency-triggered policy ($r^*=1$, $W=2$) under shock and mechanism-aware scenarios.
\State Compute $\hat S^{(1)}_i(t)$ for each regime; report $\Delta S(T_{policy})$ and $\Delta S(T_{eval\_policy})$ by scenario.
\State Compute diagnostics: $AUC_{row}$, IPCW Brier, IBS, ECE at reported horizons.
\State Compute $\bar\mu^{(a)}_g(t)$, $Gap^{(a)}(t)$, $\Delta Gap(t)$ for gender and disability subgroups.
\State Summarize subgroup uncertainty via $B=500$ bootstrap CIs.
\end{algorithmic}
\end{algorithm}

\begin{table}[tbp]
\centering
\caption{Mapping of Algorithm~\ref{alg:pipeline} steps to pipeline scripts.}
\label{tab:app_script_map}
\small
\begin{tabular}{lll}
\hline
Steps & Script(s) & Role \\
\hline
1       & \texttt{A\_01}, \texttt{A\_02}                    & Load OULAD; build enrollment backbone \\
2--4    & \texttt{A\_03}                                    & Endpoint, censoring anchor, $t^{final}$ \\
5       & \texttt{A\_04}                                    & Person-period weekly frame \\
6       & \texttt{A\_05}                                    & Weekly covariate engineering \\
7       & \texttt{A\_06}                                    & Stratified temporal split \\
8--9    & \texttt{B\_00}                                    & Hazard model, predictions \\
10--12  & \texttt{C\_00}, \texttt{C\_01}, \texttt{C\_02}   & Scenario catalog; shock and mechanism-aware; $\Delta S$ \\
13      & \texttt{D\_00}, \texttt{D\_01}                    & Survival metrics and robustness diagnostics \\
14      & \texttt{E\_01}, \texttt{E\_01b}                   & Subgroup fairness (gender; disability) \\
\hline
\end{tabular}
\end{table}

\section{Appendix C: Hyperparameter Configurations}
\label{app:c}

\paragraph*{Primary model.}
Logistic regression: \texttt{solver=liblinear}, \texttt{C=1.0}, \texttt{max\_iter=4000}, \texttt{class\_weight=balanced}, \nolinkurl{random_state=42}. Numeric features standardised; categorical variables one-hot encoded. Calibration: \texttt{CalibratedClassifierCV(method="sigmoid", ensemble=False)}, GroupKFold $k=5$ (enrollment-grouped)~\cite{Kull2017BetaCalibration}.

\paragraph*{Shock intensity scenarios.}
Three named intensities: $\delta_{shock}\in\{0.08, 0.20, 0.60\}$. Dose-response sweep: $\delta_{shock}\in\{0.02,0.04,\ldots,0.75\}$ (19 values). Sensitivity grid: 216 combinations of trigger/schedule parameters.

\paragraph*{IPCW Brier and IBS equations.}
\begin{equation*}
BS_{IPCW}(T)=\frac{1}{n}\sum_i w_i(T)\left(Y_i^{(T)}-\hat p_i^{(T)}\right)^2, \quad
IBS(0{:}T)=\frac{1}{T+1}\sum_{t=0}^{T}BS_{IPCW}(t).
\end{equation*}

\section*{Declaration of Generative AI Software Tools}

During the preparation of this work, the authors used large language model tools to assist with academic writing tasks including language refinement, wording suggestions, and structural editing. After using these tools, the authors reviewed and edited the content as needed and take full responsibility for the content of the publication. No generative AI was used to generate study data, statistical analyses, results, or conclusions.

\phantomsection
\section*{Acknowledgments}
\addcontentsline{toc}{section}{Acknowledgments}
The authors thank the Open University for making the OULAD dataset publicly available.

\section*{Declaration of Conflicting Interest}
\addcontentsline{toc}{section}{Declaration of Conflicting Interest}
The author(s) declared no potential conflicts of interest with respect to the research, authorship, and/or publication of this article.

\section*{Funding}
\addcontentsline{toc}{section}{Funding}
This research received no specific grant from any funding agency in the public, commercial, or not-for-profit sectors.

\phantomsection
\bibliography{ref}


\end{document}